\documentclass{article}

\usepackage{graphicx}
\usepackage{amsmath}
\usepackage{hyperref}
\usepackage{amssymb}

\title{Attempted Blind Constrained Descent Experiments}
\author{Prasad N R\footnote{"Relevant author's information" section has more information} \\
	Alumnus, Electrical and Computer Engineering, Carnegie Mellon University \\ praghave@alumni.cmu.edu}

\begin{document}
\maketitle
\begin{abstract}
	Blind Descent uses constrained but, guided approach to “learn” the weights; The probability density function is non-zero in the infinite space of the dimension (case in point: Gaussians and normal probability distribution functions). In Blind Descent paper, some of the implicit ideas involving layer by layer training and filter by filter training (with different batch sizes) were proposed as probable greedy solutions. The results of similar experiments are discussed. Octave (and proposed PyTorch variants') source code of the experiments of this paper can be found at \url{https://github.com/PrasadNR/Attempted-Blind-Constrained-Descent-Experiments-ABCDE-}. This is compared against the ABCDE derivatives of the original PyTorch source code of \url{https://github.com/akshat57/Blind-Descent}.
\end{abstract}

\section{Introduction}
In Blind Descent, normal probability distribution is used. This is better than zero mean unit uniform distribution which has no memory to store the mean. In zero mean unit uniform distribution, search for weights is global. In order to fully understand the effects of constraints, the range has to be restricted as well. So another uniform distribution with saved weights as mean and $[-\eta, \eta]$ range around the mean is compared with the rest. Optimisers like Adam and SGD are not used. Bias is not used in any of the layers (as that might introduce scaling issues as constrained random descent is enforced).\\

A CNN consisting of several layers with only convolutions, max-pooling and activations are considered. Fully connected dense network is replicated using $1\times1$ filters. (So, technically, MLPs are not used in the network at all even for fully connected networks) Each convolution layer is supposed to contain multiple filters. The convolution layer sums the input feature maps (or input channels) and performs the convolution with each of the filters (or kernels) to produce output feature maps (or output channels). Each output feature map is a result of convolution with each of the filters. So, the number of output channels is (equal to) the number of filters. Experiments in this paper involve ReLU as an activation layer (Rectified Linear Unit). Max-pooling is used for pooling. Both ReLU and Max-pooling preserve the number of channels. The filters are randomly initialised just like that of Blind Descent. But, unlike normal probability alone experimented with in Blind Descent paper, in this paper, three probability distributions are experimented with: Zero mean unit uniform random distribution, uniform random distribution and normal probability distribution. Both uniform random distribution and normal distribution have saved weights as mean. Uniform distribution has the range $[-\eta, \eta]$ around the mean and normal distribution has learning rate as standard deviation.\\

For these three random probability distributions, layer by layer training is performed by considering the saved network, freezing all layers except the one that is being trained. Then, similarly, this layer by layer training is applied to all layers cyclically. If the loss after applying the random probability distribution is lesser than that of the saved network, the new network is saved. This process is repeated for each batch.\\

For these three random probability distributions, random freezing is experimented with. Similar to layer by layer training, filters are frozen. But, here, the difference is in freezing a certain number of filters in the whole network randomly for each batch.\\

For these three probability distributions, PyTorch implementation is also experimented with. Batch size is varied for original Blind Descent implementation. Gradient Check training is experimented with.

\section{Prior work}
Akshat 'et alii' proposed Blind Descent which is the "backbone" of this research.\cite{gupta2020blind} Layer by layer training can be extended to Blind Descent.\cite{brock2017freezeout} In the original FreezeOut paper, the authors mention making several layers, performing inference and the layer that is trained would require backpropagation for dynamic graph framework. But, with Blind Descent, it is as easy as just updating the weights using just one call to random distribution function in one iteration without having to traverse through the previous layers (as in backpropagation). Random freezing does not work well for CNNs with backpropagtion unless momentum is used.\cite{randomFreezing} So, this method is tested with Blind Descent (with different probability distributions). There has been discussions about batch size.\cite{masters2018revisiting} So, different batch sizes are experimented with.

\section{Proposed variants}
Blind Descent follows the following update rule:
\[
x^{(t+1)} = \begin{cases} 
\mathit{d}(\mu = x^{(t)}, \sigma = \eta) , & \text{if }\mathbf{L}(x^{(t+1)}) < \mathbf{L}(x^{(t)}) \\ 
x^{(t)} & \text{otherwise} \end{cases}
\]\

This is the same as:
\[
x^{(t+1)} = \begin{cases} 
x^{(t)} + \mathit{d}(\mu = 0, \sigma = \eta) , & \text{if }\mathbf{L}(x^{(t+1)}) < \mathbf{L}(x^{(t)}) \\ 
x^{(t)} & \text{otherwise} \end{cases}
\]\

\subsection{Two more update probability distributions}
\begin{enumerate}
	\item \[
	x^{(t+1)} = \begin{cases} 
	x^{(t)} + \mathit{U}(-\eta, \eta) , & \text{if }\mathbf{L}(x^{(t+1)}) < \mathbf{L}(x^{(t)}) \\ 
	x^{(t)} & \text{otherwise} \end{cases}
	\]\
	
	Where \textit{U} is the uniform random distribution in this case with range $[-\eta, \eta]$.
	
	\item \[
	x^{(t+1)} = \begin{cases} 
	\mathit{U}(-1, 1) , & \text{if }\mathbf{L}(x^{(t+1)}) < \mathbf{L}(x^{(t)}) \\ 
	x^{(t)} & \text{otherwise} \end{cases}
	\]\
	
	Where \textit{U} is the zero mean unit uniform random distribution in this case.
	
\end{enumerate}

\subsection{Two more training methods}
\begin{enumerate}
	\item Layer by layer training.
	\[
	x^{(t+1)}.layer[i] = \begin{cases} 
	\mathit{d}(\mu = x^{(t)}.layer[i], \sigma = \eta), & \text{if }\mathbf{L}(x^{(t+1)}) < \mathbf{L}(x^{(t)}) \\ 
	x^{(t)} & \text{otherwise} \end{cases}
	\]\
	Where i = iteration mod number of layers
	\item Random freezing.\\
	Let $\gamma$ be the freeze probability. Let \textit{F} be the set of frozen weights. Let us consider $j^{th}$ filter in the network. Then,
	\[
	x^{(t+1)}_j = \begin{cases} 
	x^{(t)}_j, & \text{if } x^{(t)}_j \epsilon F \; or \; \mathbf{L}(x^{(t+1)}) > \mathbf{L}(x^{(t)}) \\ 
	\mathit{d}(\mu = x^{(t)}_j, \sigma = \eta) & \text{otherwise} \end{cases}
	\]\
	
	In both layer by layer training and random freezing, $\mathit{d}(\mu = x, \sigma = \eta)$ refers to probability distribution with saved weights as mean and learning rate as the standard deviation/range.
\end{enumerate}

\section{Octave experiments}
Several experiments were performed with Octave designed from scratch to observe the effectiveness of random probability distributions and to make sure there are no limitations due to a particular library.\cite{Octave} Cross-entropy loss is used with softmax. $\eta$ = 0.001, number of epochs = 40, image range = [0, 1], batch size = 16, $\gamma$ = 0.75. ReLU is the activation used.\\

As a preprocessing of softmax though, for each batch, the prediction values are divided by the standard deviation and all prediction values are subtracted by the maximum value of the predictions. Then, this vector is the input of softmax followed by cross entropy loss calculation.\ref{Octave}\\

\begin{table}
	\begin{tabular}{|c|c|c|c|c|} \hline
		{\textbf{\shortstack{Experiment\\number}}} & {\textbf{Guidance}} & {\textbf{Freezing}} & {\textbf{\shortstack{CIFAR-10\\test accuracy (\%)}}} & {\textbf{\shortstack{MNIST\\test accuracy (\%)}}} \\ \hline
		{one} & {\shortstack{None\\(Unit Random)}} & {None (Memoryless)} & {11.88} & {10.24} \\ \hline
		{two} & {\shortstack{None\\(Unit Random)}} & {Layer by Layer} & {10.29} & {9.74} \\ \hline
		{three} & {\shortstack{None\\(Unit Random)}} & {Random Freezing} & {11.42} & {9.8} \\ \hline
		{four} & {Uniform} & {None (Memoryless)} & {9.27} & {12.65} \\ \hline
		{five} & {Uniform} & {Layer by Layer} & {9.8} & {10.52} \\ \hline
		{six} & {Uniform} & {Random Freezing} & {11.25} & {6.9} \\ \hline
		{seven} & {Normal} & {None (Memoryless)} & {10.82} & {9.55} \\ \hline
		{eight} & {Normal} & {Layer by Layer} & {9.51} & {11.35} \\ \hline
		{nine} & {Normal} & {Random Freezing} & {10.85} & {10.17} \\ \hline
	\end{tabular}
	\caption{List of Blind Descent experiments performed with CNN in Octave}
	\label{Octave}
\end{table}

\begin{figure*}
	\includegraphics[width=\textwidth]{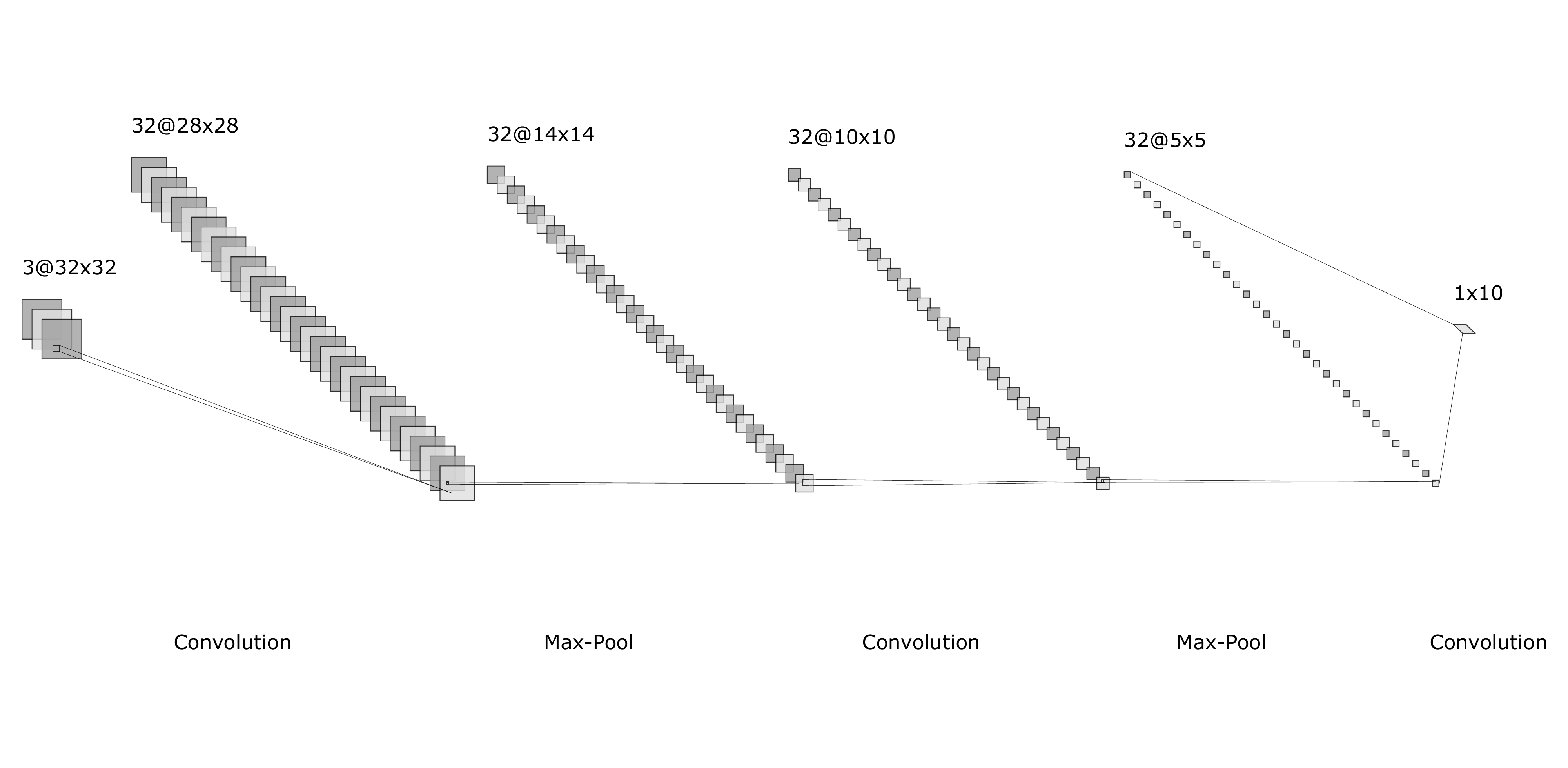}
	\caption{Architecture used for Blind Descent CNN experiments performed with Octave; This diagram was prepared using \url{http://alexlenail.me/NN-SVG/}}
\end{figure*}

A three layer CNN is used. Fully connected dense layers are not used. The forward pass looks like the following. Note that, only here in this code snippet, \textit{x} refers to the data and does not refer to weights (\textit{x} refers to the weights of the network throughout this paper, similar to the conventions used in Blind Descent paper).

\begin{footnotesize}
	\begin{verbatim}
	x = convolution(x, CNNtable, "layer1", activation = false);
	x = pooling(x);
	x = convolution(x, CNNtable, "layer2");
	x = pooling(x);
	x = convolution(x, CNNtable, "layer3", activation = false);
	predictions = reshape(x, 1, numberOfClasses);
	\end{verbatim}
\end{footnotesize}

\section{PyTorch experiments}
These experiments were performed by modifying the original Blind Descent architecture using PyTorch.\cite{gupta2020blind}\cite{pytorch} The same CNN architecture used for Octave experiments is used.

\subsection{Different batch sizes for Blind Descent}
For $\eta$ = 0.001 and 40 epochs, Blind Descent is run for different batch sizes of MNIST for two random probability distribution functions: uniform and normal.\ref{batch_size}

\begin{table*}
	\centering
	\begin{tabular}{|c|c|c|} \hline
		{\textbf{Batch size}} & {\textbf{\shortstack{MNIST (uniform)\\test accuracy (\%)}}} & {\textbf{\shortstack{MNIST (normal)\\test accuracy (\%)}}} \\ \hline
		{16} & {74.08} & {68.25} \\ \hline
		{32} & {70.42} & {67.24} \\ \hline
		{64} & {70.33} & {69.28} \\ \hline
		{128} & {72.79} & {74.61} \\ \hline
		{256} & {67.98} & {72.13} \\ \hline
		{512} & {74.74} & {67.59} \\ \hline		
	\end{tabular}
	\caption{Different batch sizes for Blind Descent}
	\label{batch_size}
\end{table*}

\subsection{Blind Descent for different probability distributions}
For $\eta$ = 0.001, batch size = 64 and 40 epochs, different probability distributions are considered for two datasets.\ref{differentProbabilityDistributions}

\begin{table*}
	\centering
	\begin{tabular}{|c|c|c|} \hline
		{\textbf{\shortstack{Probability\\distribution}}} & {\textbf{\shortstack{CIFAR-10\\test accuracy (\%)}}} & {\textbf{\shortstack{MNIST\\test accuracy (\%)}}} \\ \hline
		{Zero mean unit uniform} & {9.98} & {9.46} \\ \hline
		{Uniform} & {9.86} & {75.07} \\ \hline
		{Normal} & {10.01} & {69.09} \\ \hline
	\end{tabular}
	\caption{Different probability distributions for Blind Descent}
	\label{differentProbabilityDistributions}
\end{table*}

\subsection{Gradient Check: Attempted combination of advantages of Blind Descent and Gradient Descent}
The attempt was to learn with a fraction of dataset (Few shot or even Fraction shot learning) using constrained random learning rates (for not just first order loss derivative but, with higher order derivatives as well) with Blind Descent. The premise is to continuously guarantee lowering loss values for larger batch sizes. For each batch, the CNN filter weights are updated as follows:

\[
x^{(t+1)} = \begin{cases} 
x^{(t)} - \eta_1\frac{\partial L}{\partial x} - \eta_2\frac{\partial^2 L}{\partial x^2} - ..., & \text{if }\mathbf{L}(x^{(t+1)}) < \mathbf{L}(x^{(t)}) \\ 
x^{(t)} & \text{otherwise} \end{cases}
\]\

where $\eta_1$, $\eta_2$ ... $\epsilon$ \textit{d}(a, b) such that a is the lower limit of the probability distribution and b is the higher limit of probability distribution function \textit{d}.\\

In the experiments performed, first order i.e., $\eta = \eta_1$ is considered with $\textit{d} = 10^{U(-6,\;  1)}$ where \textit{U} is the uniform random probability distribution. For image range [0, 1], 40 epochs and a batch size of 256, the test accuracies achieved for MNIST is 97.69\% and CIFAR-10 is 41.25\%.

\section{Conclusions and Discussions}
Several experiments were performed on Blind Descent with constraints. It is not clear why PyTorch version of the Blind Descent code works with MNIST. It might be due to frequency domain FFT convolutions or due to initialisation.

\section{Relevant author's information}
I am Prasad N R (Prasad Narahari Raghavendra). I applied only once for Master's of Science, Electrical and Computer Engineering, College of Engineering, Carnegie Institute of Technology ("Carnegie Tech"), Carnegie Mellon University, Pittsburgh, PA (Pennsylvania), United States (US). I applied from Bangalore (Karnataka, India) by the end of 2018 (14 October 2018) for which I was offered admission on 4 April 2019 and I accepted it on 14 April 2019. This is my first study program after completing my Bachelor of Engineering in 2016 (after working for commercial companies for a while in between). While I applied for Fall 2019, I was offered admission for Spring 2020 "due to limited size of the M.S. class and the constraints of physical facilities" which I accepted. I traveled away from India for the first time in my life and reached US on 30 December 2019 (via Dubai, UAE). Since then, I have stayed in Pittsburgh and never traveled outside Pittsburgh. I completed usual two year Master's degree in one year. I started my degree program on 10 January 2020 and completed it on 23 December 2020 being a full-time on-campus student (I was physically attending classes till March 2020; Due to COVID-19 scenario, I opted for online (internet) classes thereafter). I received my electronic degree certificate on 12 February 2021 and physical degree certificate on 16 February 2021. The language used in the research may be related to software program code and not general English. This research does not constitute "self work" as all of these ideas were independently generated (but not validated until my graduation in 2020) while I was simultaneously completing the Master's degree.

\section{Conflicts of interests}
There are no conflicts of interests.

\end{document}